\documentclass{article}

    \usepackage[preprint]{neurips_2025}

\usepackage[utf8]{inputenc} 
\usepackage[T1]{fontenc}    
\usepackage{url}            
\usepackage{booktabs}       
\usepackage{amsfonts}       
\usepackage{nicefrac}       
\usepackage{microtype}      
\usepackage{multirow}
\usepackage{multicol}
\usepackage{amsmath}
\usepackage{wrapfig}
\usepackage{framed}
\usepackage{floatrow}
\PassOptionsToPackage{table,dvipsnames}{xcolor}
\usepackage[colorlinks=true,linkcolor=MidnightBlue,anchorcolor=black,citecolor=MidnightBlue,urlcolor=MidnightBlue, pdfborderstyle={/S/U/W 0.5}]{hyperref} 
\newcommand{\name}{\textsc{PDDL-Instruct}}
\newcommand{\nameS}{\textsc{PDDL-Instruct} }

\newcommand{\CC}{\cellcolor{gray!13}}

\usepackage[framemethod=TikZ]{mdframed}
\mdfdefinestyle{PVFrame}{%
    frametitlerule=true,
    nobreak=true,
    frametitlebackgroundcolor=black,
    frametitlefont={\bfseries\color{white}},
    linecolor=black,
    outerlinewidth=0pt,
    roundcorner=0pt,
    innertopmargin=4pt,
    innerbottommargin=4pt,
    innerrightmargin=8pt,
    innerleftmargin=8pt,
    backgroundcolor=black!5}

\newcommand{\definespeaker}[3]{%
  \newcommand{#1}[1]{\textbf{#2: }\textcolor{#3}{##1}}%
}
\definecolor{bg}{HTML}{4285F4}
\usepackage{listings}
\usepackage[most]{tcolorbox}
\newcommand{\mytcbinput}[3]{
\tcbinputlisting{
    colback=gray!5,colframe=MidnightBlue,
      listing file=#1,
      breakable,
      title=#2,
      listing only
    }
}
\usepackage{array}
\usepackage{paracol}
\usepackage{colortbl}
\usepackage{algorithm,algpseudocode}

\definespeaker{\Psays}{Pulkit}{red}
\definespeaker{\Asays}{Anthony}{blue}
\definespeaker{\Nsays}{Nicole}{violet}
\definespeaker{\Ssays}{Swaroop}{green}

\title{Teaching LLMs to Plan: Logical Chain-of-Thought Instruction Tuning for Symbolic Planning}

\author{%
  Pulkit Verma \\
  MIT CSAIL\\
  Cambridge, USA \\
  \texttt{pulkitv@mit.edu} \\
  \And
  Ngoc La \\
  MIT CSAIL\\
  Cambridge, USA \\
  \texttt{ntmla@mit.edu} \\
  \And
  Anthony Favier \\
  MIT CSAIL\\
  Cambridge, USA \\
  \texttt{antfav24@mit.edu} \\
  \AND
  Swaroop Mishra\thanks{Work done before joining Microsoft AI.} \\
  Microsoft AI \\
  Mountain View, USA \\
  \texttt{swaromishra@microsoft.com} \\
    \And
  Julie A. Shah \\
  MIT CSAIL\\
  Cambridge, USA \\
  \texttt{julie\_a\_shah@csail.mit.edu} \\
}

\begin{document}

\maketitle

\begin{abstract}
Large language models (LLMs) have demonstrated impressive capabilities across diverse tasks, yet their ability to perform structured symbolic planning remains limited, particularly in domains requiring formal representations like the Planning Domain Definition Language (PDDL).  In this paper, we present a novel instruction tuning framework, \name, designed to enhance LLMs' symbolic planning capabilities through logical chain-of-thought reasoning. Our approach focuses on teaching models to rigorously reason about action applicability, state transitions, and plan validity using explicit logical inference steps. By developing instruction prompts that guide models through the precise logical reasoning required to determine when actions can be applied in a given state, we enable LLMs to self-correct their planning processes through structured reflection.  The framework systematically builds verification skills by decomposing the planning process into explicit reasoning chains about precondition satisfaction, effect application, and invariant preservation. Experimental results on multiple planning domains show that our chain-of-thought reasoning based instruction-tuned models are significantly better at planning, achieving planning accuracy of up to 94\% on standard benchmarks, representing a 66\% absolute improvement over baseline models. This work bridges the gap between the general reasoning capabilities of LLMs and the logical precision required for automated planning, offering a promising direction for developing better AI planning systems. 

\end{abstract}

\section{Introduction}
\label{sec:introduction}

Large Language Models (LLMs) like GPT~\citep{achiam2023gpt}, Gemini~\citep{team2023gemini}, LLaMA~\citep{touvron2023llama}, etc. 
have demonstrated remarkable success across various domains including mathematics and 
coding~\citep{imani-etal-2023-mathprompter,gaur-saunshi-2023-reasoning,RomeraParedes2023MathematicalDF,ahn-etal-2024-large}. 
However, a critical gap remains in their ability to perform structured symbolic planning, a fundamental 
capability required for reliable real-world sequential decision-making systems. Recent studies have highlighted this issue that while LLMs excel at general reasoning over unstructured text, they struggle with the logical reasoning and systematic verification required for automated planning tasks~\citep{stechly2023gpt,valmeekam2023can,valmeekam2023planning,kambhampati2024position,stechly2025on}.

This limitation becomes particularly evident when considering formal planning representations such as the Planning Domain Definition Language (PDDL)~\citep{McDermott_1998_PDDL}.
Despite some promising results with specific configurations~\citep{liu2023llm,wang2023voyager}, these models generally perform poorly on multi-step reasoning tasks including classical planning~\citep{hsiao2025a}. 
This has significant implications 
for planning tasks, which are PSPACE-complete~\citep{bylander1991complexity} and inherently require scaling reasoning efforts with problem complexity.

\begin{figure*}
    \centering
    \includegraphics[width=\linewidth]{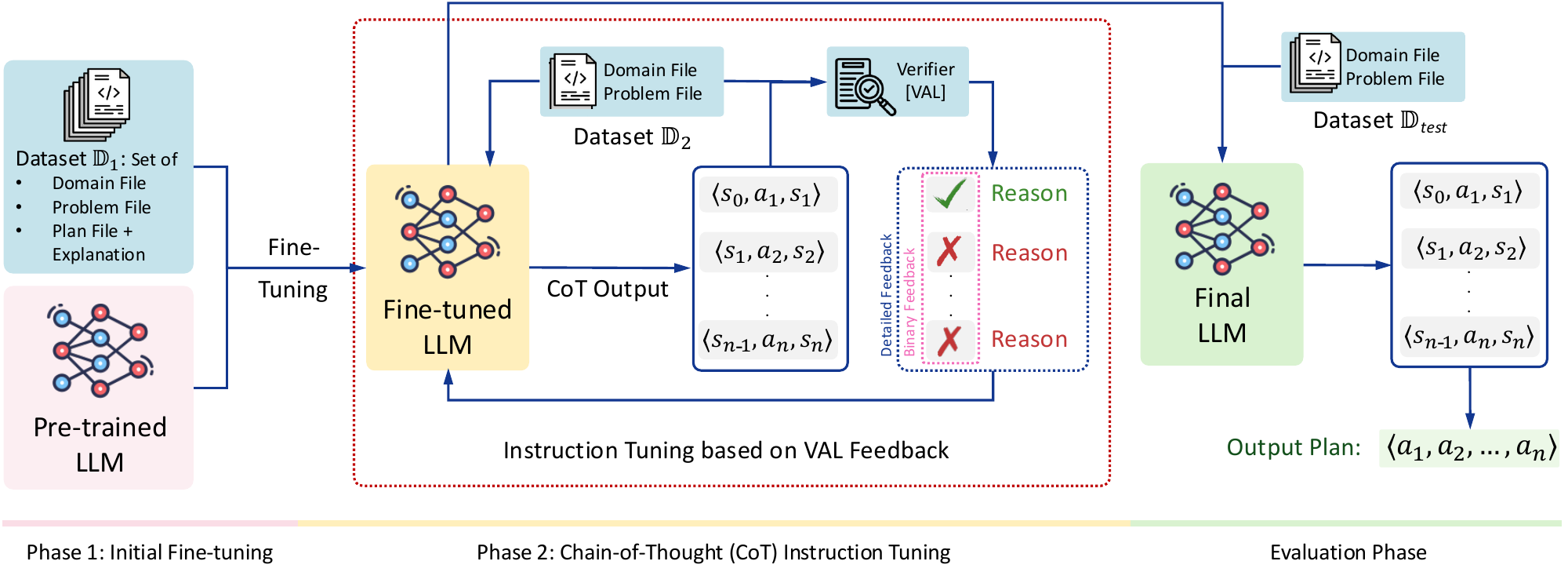}
    \caption{The \nameS approach consists of three phases: Two training phases (Initial and CoT Instruction Tuning) and evaluation phase. The main innovation lies in the second phase: CoT Instruction Tuning (highlighted by the red boundary). The initially tuned LLM is further trained using a structured instruction process that emphasizes complete logical reasoning chains.}
    \label{fig:flowchart}
\end{figure*}

In this paper, we challenge this limitation by introducing \emph{\name}, a novel framework shown in Fig.~\ref{fig:flowchart}, that augments LLMs' reasoning capabilities with the formal reasoning required for automated planning.
\nameS explicitly teaches LLMs to reason through the precondition-effect structure of planning domains using logical chain-of-thought prompting.
By decomposing planning verification into atomic reasoning steps and incorporating this structure into instruction tuning, our approach enables LLMs to not only generate syntactically correct plans but also to verify their logical validity through step-by-step reasoning. 
This ability to perform structured verification significantly enhances the reliability of LLM-generated plans and opens up possibilities for self-correction through iterative refinement. 

Main contributions of this paper are:
\begin{itemize}
    \item A novel instruction tuning framework that enhances symbolic planning capabilities in LLMs through logical chain-of-thought reasoning, focusing specifically on plan generation and action applicability verification.
    
    \item A formalization of the planning verification process as decomposable reasoning chains, enabling LLMs to systematically check preconditions, apply effects, and validate invariants.
    
    \item Empirical evidence demonstrating that instruction-tuned LLMs can develop robust planning capabilities that generalize across domains.
    
\end{itemize}

Our results show that \nameS significantly outperforms both baseline models and traditionally instruction-tuned models, achieving planning validity rates of up to 94\% in standard planning domains. This work not only addresses a critical limitation in current LLM capabilities but also provides a foundation for developing more trustworthy AI systems capable of reliable planning in complex scenarios.

\section{Related Work}
\label{sec:relatedWork}

\paragraph{LLMs for planning} 
Various approaches have been recently used for using LLMs for planning, such as generating executable code dictating the planned behaviors ~\citep{liang_code_2023, singh_progprompt_2023, nijkamp2023codegen, wang2025planning}, using closed loop with environment feedback ~\citep{huang2022inner, song2023llm} or for self-refinement ~\citep{wang2023describe, zhou2024isr}. A few recent approaches also synthesize Python programs using LLMs for planning~\citep{silver2024generalized,hao2025planning,chen2025language,correa2025classical}.

However, as summarized in~\citep{tantakoun2025llms}, LLMs face challenges with long-term planning and reasoning, often producing unreliable plans  ~\citep{stechly2024chain, pallagani2023understanding, momennejad2023evaluating}, frequently failing to account for the effects and requirements of actions as they scale ~\citep{stechly2024chain}, and their performance degrades with self-iterative feedback~\citep{stechly2023gpt, valmeekam2023can, huang2025planning}.

Another approach consists in using LLMs to generate automated planning models (e.g. PDDL domain and problem) and to rely on existing symbolic solvers to produce sound solutions. This hybrid paradigm has received a lot of interest~\citep{huang2025planning, mahdavi2024leveraging, zhang2024lamma, tantakoun2025llms}. 
Still, generating such structured models accurately is challenging for LLMs. To reach high accuracy, the process usually relies on human interventions for feedback and validation ~\citep{guan2023leveraging}, using external verifiers~\citep{silver2024generalized,hao-etal-2025-large}, or focuses on limited aspects of the problem (e.g. only generating planning goals ~\citep{xie2023translating}.
NL2P~\citep{gestrin2024towards} proposes to use explicit inference steps and Chain of Thoughts back prompting to generate the PDDL domain and problem from natural language inputs. 
Here, we propose to finetune an LLM to learn explicit inference steps to reason on action applicability, state transitions, and plan validity to generate a plan.

Finetuning for planning improves significantly the model's capabilities to generate symbolic plans~\citep{pallagani2023understanding, li-etal-2025-unlocking,fu2025agentrefine}. However, the main drawbacks of this approach are its high economic, time, and computational costs, as well as the degradation of the transferability of the model. Finetuning makes the model specialized on the domains and problem types trained on, with poor transferability to new problems.

\paragraph{Instruction tuning} Instruction tuning has emerged as a significant approach in NLP to enable zero-shot generalization on unseen tasks~\citep{mishra-etal-2022-cross,wei2022finetuned,Ouyang2022training}. This technique involves fine-tuning large language models to perform diverse tasks by following instructions, making the task source crucial for effective tuning~\citep{longpre2023flan}. While existing methods often rely on human-crowdsourced tasks from datasets like T0~\citep{sanh2022multitask}, FLAN~\citep{wei2022finetuned,longpre2023flan}, and NaturalInstructions~\citep{mishra-etal-2022-cross,wang-etal-2022-super}, these high-quality resources demand significant human effort and are typically limited in quantity. An alternative approach involves model-generated tasks, where powerful language models like GPT-3 and GPT-4 generate diverse instructions and task pairs~\citep{wang-etal-2022-super,peng2023instruction}, though these can introduce noise when outputs don't properly correspond to inputs. In this work, we alleviate this problem by leveraging the automated planning task generators~\citep{seipp-et-al-zenodo2022,valmeekam2023planbench} to create the instruction tuning dataset.

\paragraph{Chain-of-Thought Reasoning} A significant advancement in improving LLM reasoning ability is the implementation of Chain of Thought (CoT) prompting~\citep{wei2022CoT}. By generating explicit intermediate reasoning steps, these models can now address complex logical deduction and multistep problem-solving. Short CoT approaches~\citep{lambert2024t,kojima2022large} demonstrated effectiveness for straightforward problems but revealed limitations when confronting more intricate challenges. The evolution toward longer reasoning chains has subsequently transformed the landscape of machine reasoning. \citet{stechly2024chain} argued that despite its efficacy for reasoning tasks, CoT is not suitable for planning, but in this work we show that with proper integration of instruction tuning using better prompts, CoT can indeed be used for planning tasks.

\section{Preliminaries}
\label{sec:preliminaries}

\paragraph{Automated Planning}
In this section, we briefly describe automated planning. Please refer to \citet{geffner2013planning} and \citet{chen2025aiplanning} for more details.

An automated planning problem can be formally characterized as a tuple $\langle P, A, s_0, G \rangle$, where $P$ is a set of fluents used to describe a discrete and fully-observable state $S$, $A$ represents a finite set of actions, $s_0 \in S$ denotes the initial state, and $G$ specifies the goal conditions. Each action $a_i \in A$ is defined as $\langle pre(a_i), add(a_i), del(a_i)\rangle$, where $pre(a_i)$ is the set of fluents that must hold in the current state for the action to be executable, $add(a_i)$ is the set of fluents that become true after executing $a_i$, and $del(a_i)$ is the set of fluents that become false after executing $a_i$. Note that the state space $S$ in classical planning emerges from all possible truth assignments to the set of fluents. 

A solution to a planning problem $\mathcal{P}$, called a plan $\pi$, is a sequence of actions $\langle a_0, a_1, ..., a_{n-1}\rangle$ that transforms the initial state into one satisfying the goal conditions after $n$ steps. Note that $\pi$ produces state transitions $s_{i+1} = a_i(s_i) = (s_i \setminus del(a_i)) \cup add(a_i)$ for all $0 \leq i < n$ such that $s_n \in G$.  $\pi$ is considered \emph{optimal} if it takes the least number of actions (in this work, we consider actions with uniform cost) to reach a goal state, whereas it is considered \emph{satisficing} if it reaches the goal successfully but with more actions than needed by an optimal plan.

The Planning Domain Definition Language (PDDL)~\citep{McDermott_1998_PDDL}, based on STRIPS~\citep{Fikes1971}, provides a standardized specification for automated planning problems. PDDL consists of a \emph{domain} $\mathcal{D}=\langle P,A\rangle$ containing the sets of fluents $P$ and actions $A$ (along with their precondition, $add$ and $del$ sets), and a \emph{problem} $\mathcal{P} = \langle s_0,G\rangle$ containing the initial state $s_0$, and a goal condition $G$.

\paragraph{Instruction Tuning} Instruction tuning~\citep{mishra-etal-2022-cross,wei2022finetuned,Ouyang2022training} is an approach for fine-tuning LLMs on a labeled dataset.
Consider an instruction tuning dataset $\mathbb{D}_{\text{1}} = \{(x_i, r_i)\}_{i=1}^\Omega$ with $\Omega$ labeled samples, where $x_i$ represents an instruction and $r_i$ its corresponding ideal response. We denote our large language model as $\mathcal{M}_\theta$ with parameters $\theta$. The model produces output $y_i = \mathcal{M}_\theta(x_i)$ for a given instruction $x_i$. The standard instruction tuning objective aims to find model parameters $\theta^*$ that minimize expected discrepancy (loss $\mathcal{L}$) between model predictions ($\mathcal{M}_\theta(x)$) and target responses ($\tau$) across the instruction dataset (Dataset $\mathbb{D}_1$, as described in Sec. \ref{sec:problem}):
\begin{equation}
\theta^* = \arg\min_\theta \mathbb{E}_{(x,\tau) \sim \mathbb{D}_{\text{1}}} [\mathcal{L}(\mathcal{M}_\theta(x), \tau)]
\end{equation}

\paragraph{Chain-of-thought reasoning} 

Chain-of-Thought (CoT) reasoning can be formally defined as a structured decomposition of a complex reasoning task into an explicit sequence of intermediate logical steps.
Given a problem input $x$ and a target output $y$, a chain-of-thought reasoning process $\mathcal{R}$ is a sequence of $K$ intermediate reasoning states
$\mathcal{Z}(x) = (z_1, z_2, \ldots, z_K)$, where each $z_i$ represents an atomic reasoning step that transforms the latent state from $z_{i-1}$ to $z_i$, with $z_0$ implicitly defined as the initial problem state derived from $x$.
Each reasoning step $z_i$ can be characterized as a tuple $z_i = (s_i, j_i, u_i)$, 
where
$s_i$ represents the symbolic state (the set of derived facts or assertions at step $i$), 
$j_i$ represents the justification (the logical rule or inference applied), and
$u_i$ represents the uncertainty estimate (the model's confidence in this reasoning step). For simplicity, going forward we will use symbolic states $s_i$ to represent reasoning states $z_i$, when clear from context, as they have a one-to-one mapping for this work. We also do not use $u_i$ estimates for this work, and the LLM is directly asked for the resulting symbolic states in each CoT step.

Two important properties that characterize effective chain-of-thought reasoning are: (i) logical coherence~\citep{wei2022CoT}, and (ii) progressive refinement~\citep{du2025think}.
A CoT process $\mathcal{R}(x)$ exhibits \emph{logical coherence} if for each step $z_i$ with $i > 1$, $\exists j_{i-1} \text{ such that } j_{i-1}(s_{i-1}) \Rightarrow s_i$,
meaning each state follows logically from the application of a justifiable inference rule to the previous state. A CoT process $\mathcal{R}(x)$ exhibits \emph{progressive refinement} if $I(z_i; y) > I(z_{i-1}; y) \quad \forall i \in \{1,2,...,K\}$,
where $I(z_i; y)$ represents the mutual information between reasoning state $z_i$ and the target output $y$.

\section{Problem Formulation}
\label{sec:problem}

\paragraph{Input} In this work, we use the following inputs: (i) a pre-trained LLM $\mathcal{M}$ as input, (ii) a dataset $\mathbb{D}$ of planning domains and problems expressed in PDDL with their solutions (satisficing plans), and (iii) a plan validator $\mathcal{V}$ used to validate the correctness of plans generated by $\mathcal{M}$. The dataset $\mathbb{D}$ consists of:

\begin{enumerate}
    \item A set $\{\mathcal{D}_1, \mathcal{D}_2, ..., \mathcal{D}_n\}$ of planning domains expressed in PDDL. 
    \item For each domain $\mathcal{D}_i$, we have problems $\mathbb{P}_i = \{\mathcal{P}_{i,1}, \mathcal{P}_{i,2}, ..., \mathcal{P}_{i,m_i}\}$. 
    \item For each planning problem $P_{i,j}$, we have a mix of valid and invalid plans $\Pi_{i,j} = \{\pi_{i,j,1}, \pi_{i,j,2}, ..., \pi_{i,j,k_{i,j}}\}$, where each plan $\pi_{i,j,l}$ is a sequence of grounded actions; and their corresponding explanations of their correctness or errors, as needed.
\end{enumerate}

\paragraph{Data Splitting}

As shown in Fig.~\ref{alg:pddl_instruct}, our approach has three phases (more details in Sec.~\ref{sec:methodology}).  To facilitate this, we partition the dataset $\mathbb{D}$ into three sets: $\mathbb{D}_{\text{1}}$, $\mathbb{D}_{\text{2}}$, and $\mathbb{D}_{\text{test}}$ for Phase 1 training, Phase 2 training, and evaluation, respectively.
We add additional data to $\mathbb{D}_{\text{1}}$ by adding incorrect plans for each problem, similar to NaturalInstructions framework~\citep{mishra-etal-2022-cross,wang-etal-2022-super}.

\paragraph{Output}

The primary output is an instruction-tuned model $\mathcal{M}_{\theta^*}$ with enhanced symbolic planning capabilities.
The model should demonstrate improved domain representation, problem representation, plan generation, action verification, plan verification, and reasoning transparency.

\paragraph{Assumptions}

Our framework assumes the planning domains follow the features explained in Sec.~\ref{sec:preliminaries}, i.e., does not contain complex PDDL features such as, e.g., conditional effects or durative actions. This simplifies the reasoning chain.

\section{\textsc{\name}: Methodology}
\label{sec:methodology}

Fig.~\ref{fig:flowchart} illustrates our comprehensive framework for enhancing symbolic planning capabilities in Large Language Models (LLMs) through logical Chain-of-Thought (CoT) instruction tuning. The approach consists of two training phases: Initial Instruction Tuning and CoT Instruction Tuning.

\subsection{Training the Model}

\paragraph{[Phase 1] Initial Instruction Tuning Phase }
In the initial instruction tuning phase (distinct from simple finetuning), we take a pre-trained LLM and train it with carefully crafted prompts that pair planning domains and problems with detailed explanations of their solutions, all derived from Dataset $\mathbb{D}_1$. As shown in Fig.~\ref{fig:flowchart}, rather than simply exposing the model to planning examples, we explicitly instruct it to analyze why each action in a plan is valid by explaining precondition satisfaction and effect application. 

This phase incorporates both correct plans and deliberately incorrect plans to teach the model to recognize and explain various planning errors. For incorrect plans, we include examples where: (1) action preconditions are not satisfied, (2) effects are incorrectly applied, (3) frame axioms are violated, or (4) the plan fails to reach the goal state. By exposing the model to both successful and failed planning attempts with detailed explanations, we establish a foundation for logical verification.

This phase establishes a foundation of planning knowledge while simultaneously teaching the model to articulate logical justifications for action validity, setting the stage for more advanced reasoning in subsequent phases. Exact prompts used in this work are available in the supplementary material.

\paragraph{[Phase 2] CoT Instruction Tuning Phase}

The main innovation of our approach lies in the CoT Instruction Tuning phase (highlighted by the red boundary in Fig.~\ref{fig:flowchart}). This second phase is itself a two-stage process described thoroughly in the next section. At a high level, in this phase, the initially tuned LLM is further trained using a structured instruction process that emphasizes complete logical reasoning chains. When presented with a domain and problem from Dataset $\mathbb{D}_2$, this initially tuned model produces step-by-step state-action-state sequences $\langle s_0, a_1, s_1\rangle, \langle s_1, a_2, s_2\rangle, \ldots, \langle s_{n-1}, a_n, s_n\rangle$ that represent a candidate plan. 

These reasoning chains are then passed through a verification module implemented using VAL~\citep{howey2004val} that systematically checks the validity of each state transition based on action preconditions and effects. Please note that while some approaches have tried using LLMs themselves as verifiers, research shows that currently LLMs do not possess sufficient self-correction capabilities in terms of reasoning~\citep{huang2024large,stechly2025on}. Unlike self-reflection approaches where models attempt to critique their own reasoning without external validation, our chain-of-thought method explicitly decomposes the planning process into verifiable logical steps, with external verification providing ground-truth feedback. This combination of explicit reasoning decomposition with verified feedback creates a more reliable foundation for enhancing planning capabilities than relying solely on the model's internal reasoning.

We explore two distinct types of verification feedback: (1) \textit{binary feedback}, which simply indicates whether an action is valid or invalid, and (2) \textit{detailed feedback}, which provides specific reasoning about each action generated by VAL in terms of which preconditions failed or which effects were incorrectly applied. Our hypothesis is that detailed feedback will lead to more robust planning capabilities by providing explicit guidance on the logical errors in the reasoning process.

The verification results provide crucial feedback that guides further instruction tuning. This feedback loop ensures that the model learns not only to generate syntactically correct plans but also to reason about their logical validity. We limit the number of times this feedback loop is used to generate new CoT plans, denoted by $\eta$. $\eta$ is a hyperparameter which we can vary to see how it affects  accuracy.

Our \nameS approach prioritizes \emph{logical coherence} (see Sec.~\ref{sec:preliminaries}) through its explicit verification of preconditions and effects at each planning step. The verification feedback ensures that each state transition follows logically from the application of a valid action, maintaining strict adherence to the domain rules. However, our approach does not ensure \emph{progressive refinement} (see Sec.~\ref{sec:preliminaries}).
This is because rather than optimizing for the shortest or most efficient plan (which would increase mutual information with an optimal solution at each step), we focus on producing satisficing plans that achieve the goal regardless of path length. Generating optimal solutions is a significantly more difficult problem in practice, both for classical planners and for training LLMs to produce them~\citep{ray2008complexity, domshlak2013complexity}.

\subsection{Training Methodology for Phase 2 CoT Instruction Tuning: Optimization Process}
A distinctive feature of our \nameS framework is the two-stage optimization process as part of the CoT Instruction Tuning that explicitly targets both the quality of logical reasoning for CoT and the resulting final planning performance. This approach addresses the unique challenges of symbolic planning by ensuring that the model not only produces correct plans but also develops robust verification capabilities through logical chain-of-thought reasoning. An algorithm for this is available in the supplementary material.

\paragraph{Stage 1: Reasoning Chain Optimization}
In the first stage, we optimize the model parameters $\theta_t$ to improve the generation of high-quality reasoning chains. Specifically, the model weight in each reasoning step $r$, $\theta^r_t$ where $t \in [0,\eta-1]$, is updated as Equation \ref{eq:reasoning}:
\begin{equation}
\label{eq:reasoning}
\theta_t^r = \theta_t - \delta_1 \nabla_{\theta_t} \mathcal{L}_{\text{reasoning}}(\theta_t, \mathbb{D}_{\text{reasoning}}^t)
\end{equation}
where $\mathcal{L}_{\text{reasoning}}$ is a loss function that measures the quality of the generated reasoning chains compared to ideal logical inference sequences, $\delta_1$ is the learning rate for this stage, and $\mathbb{D}^t_{\text{reasoning}}$ is the dataset of individual $\langle s_{i-1}, a_i, s_i \rangle$ triplets along with VAL feedback for them. This objective encourages the model to produce step-by-step reasoning that correctly (i) checks all necessary preconditions before applying actions; (ii) tracks state changes resulting from action effects; (iii) verifies that invariants are maintained throughout the plan; and (iv) detects logical inconsistencies in proposed plans.

The reasoning loss explicitly penalizes logical errors such as applying actions with unsatisfied preconditions, failing to properly propagate effects, or generating steps that violate domain constraints. By focusing specifically on the reasoning process, this stage helps the model develop the logical foundation necessary for robust planning.

\paragraph{Stage 2: End-Task Performance Optimization}
In the second stage, we optimize from the reasoning-improved parameters $\theta_t^r$ to enhance overall planning:
\begin{equation}
\theta_{t+1} = \theta_t^r - \delta_2 \nabla_{\theta_t^r} \mathcal{L}_{\text{final}}(\theta_t^r, \mathbb{D}_{\text{final}}^t)
\end{equation}
where $\mathcal{L}_{\text{final}}$ measures how well the final outputs match the expected answers in the training data, $\delta_2$ is the learning rate for this stage, and $\mathbb{D}_{\text{final}}^t$ final contains the domain, problem, and plan extracted from CoT output along with VAL feedback specifying if the plan is correct for that problem or not.
This second stage ensures that improvements in logical reasoning translate to practical planning capability of producing accurate plans.

This two-stage approach is important as Stage 1 develops the logical foundation needed for planning, while Stage 2 ensures these capabilities are properly applied to generate correct plans. The separation of these objectives allows our framework to balance between teaching fundamental reasoning skills and optimizing for task-specific performance, resulting in models that not only produce correct plans but can also reason about their correctness through explicit logical CoT inference. The exact formulations of the loss functions $\mathcal{L}_{\text{reasoning}}$ and $\mathcal{L}_{\text{final}}$ and the specific values of the hyperparameters are discussed in detail in the supplementary material.

\subsection{Evaluation Phase}

After completing both the Initial Instruction Tuning and CoT Instruction Tuning phases, the final model is evaluated in the Evaluation Phase (represented on the right side of Fig.~\ref{fig:flowchart}). In this phase, the instruction-tuned LLM is presented with new, unseen planning domains and problems from $\mathbb{D}_{test}$.
The model directly generates complete state-action-state sequences $\langle s_0, a_1, s_1\rangle, \ldots,\langle s_{n-1}, a_n, s_n\rangle$ that constitute its proposed solution to the planning problem.
These generated plans are then evaluated for correctness using VAL, but only for assessment purposes, i.e., no feedback is returned to the model. The plan is considered valid if and only if all actions in the sequence are applicable in their respective states and the final state satisfies all goal conditions.

\section{Empirical Evaluation}
\label{sec:experiments}
We conduct a comprehensive empirical evaluation of \nameS to assess its effectiveness in enhancing symbolic planning capabilities in LLMs. Our evaluation leverages PlanBench~\citep{valmeekam2023planbench}, a standardized benchmark framework for evaluating LLM planning capabilities.

We evaluate \nameS using PlanBench to assess its effectiveness in enhancing symbolic planning capabilities in LLMs. Our experiments aim to answer the following research questions:

\textbf{RQ1:} Does logical CoT instruction tuning improve plan validity compared to standard approaches?

\textbf{RQ2:} How does the quality of feedback (binary vs. detailed) affect planning performance?

\textbf{RQ3:} How well does the approach generalize across different planning domains?

We implement \nameS using Llama-3-8B and GPT-4\!~\footnote{Note that GPT-4 experiments were constrained by limited access.} foundation models.  We compare against baseline (unmodified models) and post phase 1 versions (instruction tuned on planning examples with reasoning of why each plan is valid or invalid). For \name, we test variants with binary feedback (valid/invalid) and detailed feedback (specific reasoning errors generated by VAL), each with the number of feedback iteration loop limit to $\eta \in \{10, 15\}$. All experiments were conducted on 2 NVIDIA RTX 3080 GPUs. 

\paragraph{Domains and Problems}
PlanBench provides a systematic methodology for evaluating planning capabilities across diverse planning domains and problem complexities.
We evaluate across three distinct planning domains from PlanBench, each presenting different reasoning challenges:

\begin{itemize}
   \item \textbf{Blocksworld}: The classical planning domain with blocks that can be stacked on a table or on other blocks. This domain primarily tests reasoning with a relatively small action set.
   
   \item \textbf{Mystery Blocksworld}: A more complex variant of Blocksworld with predicates identical but semantically obfuscated names.
   
   \item \textbf{Logistics}: A transportation planning domain where packages must be moved between locations using trucks and airplanes, testing the model's ability to reason about location connectivity and multi-step transport operations.
\end{itemize}

\vspace{-0.1in}
\paragraph{Evaluation Metrics}
Our primary evaluation metric is the Plan Accuracy, measuring the percentage of planning tasks for which the model generates a valid plan that achieves the specified goal. A plan is considered valid only if all actions are applicable in their respective states and the final state satisfies all goal conditions, as verified by VAL. For each domain, we generate 100 test tasks of varying complexity, with problems including different numbers of objects and requiring different plan lengths to solve.

\begin{table*}[t]
    \centering
    \small
    \begin{tabular}{l l c c c c c c c}
    \toprule
        \multirow{3}{*}{\textbf{Model}} & \multirow{3}{*}{\textbf{Domain}} & \multirow{3}{*}{\textbf{Baseline}} & \multirow{3}{*}{\textbf{Only P1}} & {\textbf{Only P2}}  & \multicolumn{4}{c}{
        \centering
        \textbf{\name}} \\ \cmidrule(lr){5-5} \cmidrule(lr){6-9}
        & & & & \textbf{Detailed} & \multicolumn{2}{c}{\textbf{Binary}}  & \multicolumn{2}{c}{\textbf{Detailed}} \\
        \cmidrule(lr){5-5} \cmidrule(lr){6-7} \cmidrule(lr){8-9}
        &          &  & & $\eta = 15$ & $\eta = 10$ & $\eta = 15$ &  $\eta = 10$ &  $\eta = 15$    \\
        \midrule
\multirow{3}{*}{Llama-3} & Blocksworld & 28\% & 78\% & 72\% & 84\% & 89\% & 91\% & 94\% \\
& Mystery BW \CC & 1\% \CC & 32\% \CC & 17\% \CC & 47\% \CC & 49\% \CC & 59\% \CC & 64\% \CC\\
& Logistics & 11\% & 23\% & 45\% & 61\% & 72\% & 75\% & 79\% \\
\midrule
\multirow{3}{*}{GPT-4} & Blocksworld \CC & 35\% \CC & 41\% \CC & 76\% \CC & 79\% \CC & 84\% \CC & 87\% \CC & 91\% \CC \\
& Mystery BW & 3\% & 17\% & 19\% & 39\% & 44\% & 54\% & 59\% \\
& Logistics \CC & 6\% \CC & 27\% \CC & 51\% \CC & 64\% \CC & 69\% \CC & 72\% \CC & 78\% \CC \\
        \bottomrule
    \end{tabular}
    \caption{Results for plan accuracy generated for 100 test tasks from each domain. Our approach \nameS was evaluated with either binary or detailed feedback. Ablation results are for only Phase 1 (P1), and only Phase 2 (P2) with detailed feedback (as it had the best performance). }
    \label{tab:main_results}
\end{table*}
\vspace{-0.1in}

\section{Results and Discussion}
\label{sec:discussion}

\paragraph{Overall Performance (RQ1)}

Tab.~\ref{tab:main_results} presents the plan accuracy across models, domains, and approaches. The results clearly demonstrate that \nameS significantly outperforms baseline models, models after Phase 1 instruction tuning, and models with just Phase 2 CoT instruction tuning.

For Llama-3, \nameS with detailed feedback and $\eta=15$ achieves validity rates of $94\%$, $64\%$, and $79\%$, respectively in Blocksworld, Mystery Blocksworld, and Logistics. 
This represents an average absolute improvement of $35\% (SD = 20\%)$ over basic instruction tuning, and of $66\% (SD = 3\%)$ over the baseline. 
Similarly, for GPT-4, \nameS with detailed feedback and $\eta=15$ achieves validity rates of $91\%$, $59\%$, and $78\%$ across the three domains.
This represents an average absolute improvement of $48\% (SD = 5\%)$ over basic instruction tuning, and of $61\% (SD = 9\%)$ over the baseline. 
These results show that logical CoT instruction tuning enhances plan accuracy significantly, not only when compared to unmodified foundation models and but more importantly, also when compared to models with only basic instruction tuning. The explicit reasoning about preconditions, effects, and state transitions enables the models to generate accurate plans.

\paragraph{Impact of Feedback Type (RQ2)}

Comparing the binary feedback and detailed feedback columns in Tab.~\ref{tab:main_results}, we observe that detailed feedback consistently outperforms binary feedback across all domains and models. For Llama-3 with $\eta=15$, detailed feedback improves plan accuracy by 5 percentage points in Blocksworld, 15 percentage points in Mystery Blocksworld, and 7 percentage points in Logistics compared to binary feedback. Note that our training approach, though developed independently, has resemblance with LEPA~\citep{zhang2025learning}, which also show that providing specific feedback about why each action fails helps in improving the reasoning capabilities of LLMs.

This pattern confirms our hypothesis that providing specific reasoning errors helps the model develop more robust verification capabilities. The advantage of detailed feedback is particularly pronounced in Mystery Blocksworld, the most complex domain with obfuscated predicates.
Additionally, we observe that increasing the iteration limit from $\eta=10$ to $\eta=15$ yields consistent improvements across all configurations. This observation indicates that the model may converge on valid plans given additional feedback iteration loops, though future experiments with varying $\eta$ are needed to confirm this. The improvement is more substantial with detailed feedback (averaging 4.3 percentage points across all domains and models) than with binary feedback (averaging 3.3 percentage points), indicating that detailed feedback enables more effective use of additional reasoning iterations. 

\paragraph{Cross-Domain Generalization (RQ3)}

Our results demonstrate significant variations in performance across domains, reflecting their inherent complexity and reasoning challenges. Both models achieve the highest performance on Blocksworld, followed by Logistics, with Mystery Blocksworld proving the most challenging.
For Llama-3 with detailed feedback and $\eta=15$, the validity rates are 94\% for Blocksworld, 79\% for Logistics, and 64\% for Mystery Blocksworld. This pattern is consistent across all configurations and models, highlighting the increasing difficulty of domains with hidden predicates and complex state interactions.
Notably, while absolute performance varies across domains, the relative improvement from \nameS is substantial in all three domains. This suggests that our approach enhances planning capabilities in a domain-general manner, with the logical reasoning framework transferring effectively across different planning scenarios.

The largest relative improvements occur in domains where baseline performance is weakest. For example, Llama-3's performance on Mystery Blocksworld improves from just 1\% to 64\% with \nameS (detailed feedback, $\eta=15$), representing a 64× improvement. This dramatic enhancement in the most challenging domain demonstrates that explicit logical reasoning is particularly valuable for complex planning scenarios where simple pattern matching is insufficient.

\section{Conclusion}
\label{sec:conclusion}

We have presented \name, a novel framework that significantly enhances the symbolic planning capabilities of Large Language Models through logical chain-of-thought instruction tuning. By decomposing the planning process into verifiable logical reasoning chains and providing explicit verification feedback, our approach enables LLMs to generate valid plans with unprecedented reliability across diverse planning domains. While our results are promising, we note that our approach does not achieve 100\% accuracy across all domains. However, when combined with frameworks like LLM-Modulo~\citep{kambhampati2024position}, which provides efficient mechanisms for integrating external tools with LLMs, our method could significantly reduce the number of required feedback loops with the verifier. 
This integration would make the planning process more efficient by allowing the model to leverage its enhanced reasoning capabilities while still benefiting from formal verification when needed, ultimately resulting in faster and more reliable planning.

A notable advantage of our VAL-based verification approach is its robustness against unfaithful chain-of-thought reasoning as described by \citet{lyu2023faithful}. While traditional CoT methods can generate plausible-sounding but internally inconsistent reasoning chains, our external verification ensures that each logical step is formally validated against the planning domain's constraints.

\paragraph{Limitations and Future Work}

While our results highlight the effectiveness of combining logical chain-of-thought with verification-guided feedback, several promising directions remain for future:

\textit{Optimizing instruction tuning data}: We can further refine our approach by applying instruction optimization techniques as described in \citet{lee-etal-2024-instruction-matters} to identify the most effective subset of instruction examples. Determining which planning scenarios and error types provide the most informative learning signal could significantly improve training efficiency.

\textit{Experimenting with more Models}: While our current evaluation across Llama-3-8B and GPT-4 demonstrates consistent improvements across distinct model paradigms and provides strong evidence for our framework's effectiveness, future work could explore additional architectures to further validate the generalizability of our approach. The consistent performance gains observed across these different model families suggest that our methodology is architecture-agnostic, though broader evaluation remains a natural extension.

\textit{Advancing to Optimal Planning}: Our current work focuses on satisficing planning—finding any valid plan that achieves the goal. A natural extension would be to incorporate plan quality metrics and develop instruction tuning approaches that guide models toward generating not just valid plans but optimal ones with minimal actions or resource usage.

\textit{Expanding PDDL Coverage}: To simplify the logical reasoning effort, we currently limit to use only a subset of PDDL features. Future work could address this limitation and incorporate more advanced PDDL features such as conditional effects, derived predicates, action costs, and temporal constraints, gradually expanding the expressive power of the planning capabilities.

\textit{Self-Verification Capabilities}: While we currently rely on an external verifier (VAL), an intriguing direction is developing self-verification capabilities where models learn to accurately critique their own plans. As LLMs continue to improve, reducing or eliminating dependence on external verifiers could make planning more autonomous and efficient.

\textit{Dynamic Iteration Control}: Our current approach uses fixed iteration limits ($\eta$). Developing techniques to dynamically determine the optimal number of iterations based on problem complexity or feedback patterns could improve efficiency, especially as we hypothesize that return will diminish on increasing $\eta$ beyond certain values.

\textit{Expanding Domain Coverage}: Currently PlanBench supports 3 domains we used in this work. Extending the evaluation to include a wider variety of planning domains would enable more comprehensive evaluation and potentially reveal new opportunities for improving logical reasoning in planning.

\textit{Beyond Planning}: Finally, the logical reasoning framework developed in this work could extend beyond planning to other sequential decision-making tasks that require long-horizon reasoning, such as theorem proving, complex puzzle solving, and multi-step logical deduction. The combination of chain-of-thought reasoning with verification-guided feedback appears to be a powerful paradigm that could enhance LLM capabilities across diverse reasoning tasks.

\section{Broader Impacts}
\label{sec:impacts}
A key positive impact is the potential to improve autonomous decision-making and to be highly beneficial to  domains such as healthcare robotics, autonomous vehicles, or disaster response. By enabling LLMs to reason about action applicability, state transitions, and plan validity, our approach supports more interpretable and verifiable AI behavior. Additionally, it contributes to bridging neural and symbolic AI, potentially democratizing access to formal planning tools for non-expert users.

However, the approach also raises risks. Over-reliance on LLM-generated plans in safety-critical domains may lead to failures. The hybrid nature of neural-symbolic reasoning may obscure responsibility and complicate error attribution. Additionally, enhanced planning capabilities could be misused for strategic manipulation or multi-step malicious behavior. To mitigate these risks, we recommend incorporating external verification, human oversight, and usage safeguards in real-world deployments.

\section*{Acknowledgments}
This work was supported in part by the ONR under grant N000142312883.

{
\bibliographystyle{plainnat}
\bibliography{llm_planning}
}

\newpage
\appendix

\section{Detailed Experimental Setup}

\subsection{Hyperparameter Configuration}
\label{sec:hyperparams}

Tab.~\ref{tab:hyperparameters} provides the complete hyperparameter configuration used in our experiments.

\begin{table}[h]
\centering
\rowcolors{2}{gray!13}{}
\caption{Complete hyperparameter configuration for \name}
\label{tab:hyperparameters}
\begin{tabular}{lcc}
\toprule
\textbf{Parameter} & \textbf{Phase 1} & \textbf{Phase 2 (CoT)} \\
\midrule
Learning Rate & 2e-5 & $\delta_1$: 1e-5, $\delta_2$: 5e-6 \\
Batch Size & 16 & 8 \\
Max Sequence Length & 2048 & 4096 \\
Training Epochs & 5 & 3 \\
Warmup Steps & 500 & 200 \\
Weight Decay & 0.01 & 0.001 \\
Gradient Clipping & 1.0 & 0.5 \\
Temperature (Generation) & 0.7 & 0.3 \\
Max Generation Length & 1024 & 2048 \\
Optimizer & AdamW & AdamW \\
$\beta_1, \beta_2$ & 0.9, 0.999 & 0.9, 0.999 \\
$\epsilon$ & 1e-8 & 1e-8 \\
Iteration Limit ($\eta$) & N/A & 10, 15 \\
\bottomrule
\end{tabular}
\end{table}

\paragraph{Learning Rates ($\delta_1$, $\delta_2$)} The learning rates control how aggressively the model weights are updated during training, with Phase 1 using a single learning rate and Phase 2 employing two distinct learning rates for its two-stage optimization process. Phase 1 uses a learning rate of $2\times10^{-5}$ for initial instruction tuning, set relatively higher because the model must learn entirely new planning capabilities from its pre-trained foundation, applying this rate to the standard cross-entropy loss when learning to generate plans with detailed explanations of action validity. Phase 2 employs two separate learning rates within its chain-of-thought instruction tuning: $\delta_1 = 1\times10^{-5}$ for Stage 1 reasoning chain optimization (Equation 2) and $\delta_2 = 5\times10^{-6}$ for Stage 2 final performance optimization (Equation 3). The first learning rate $\delta_1$ focuses on improving the quality of step-by-step logical reasoning chains, while the second learning rate $\delta_2$ is set lower to carefully optimize overall planning performance without disrupting the reasoning capabilities developed in Stage 1. Both Phase 2 learning rates are deliberately lower than Phase 1 to enable fine-tuning of the chain-of-thought reasoning without disrupting the foundational planning knowledge already acquired.

\paragraph{Batch Size} The batch size determines how many training examples are processed simultaneously before updating model weights, with values carefully chosen to balance computational efficiency with memory constraints and training dynamics. Phase 1 uses a batch size of 16, which provides sufficient gradient signal for learning basic planning concepts while remaining within GPU memory limits for the 2048-token sequences typical of initial instruction examples. Phase 2 reduces the batch size to 8 to accommodate the significantly longer chain-of-thought sequences and the additional memory overhead introduced by VAL feedback processing. The smaller batch size in Phase 2 also enables more frequent weight updates during the iterative refinement process, which is crucial for the feedback-driven learning mechanism where the model must quickly adapt to validation signals from the external verifier.

\paragraph{Maximum Sequence Length} The maximum sequence length defines the upper limit of tokens the model can process in both input and output, with values scaled to accommodate the increasing complexity of reasoning required across training phases. Phase 1 sets this limit to 2048 tokens, which sufficiently captures domain definitions, problem statements, generated plans, and basic explanations of action validity without excessive computational overhead. Phase 2 doubles this limit to 4096 tokens to accommodate the detailed chain-of-thought reasoning sequences that include comprehensive state analysis, action selection justification, explicit precondition checking, effect application reasoning, state transition tracking, and goal progress evaluation. This increased capacity is essential for the model to generate the verbose logical reasoning chains that characterize effective planning verification.

\paragraph{Training Epochs} The number of training epochs represents complete passes through the respective training datasets, with values chosen to ensure adequate learning while preventing overfitting to domain-specific patterns. Phase 1 employs 5 epochs to establish foundational planning knowledge, requiring more iterations because the model must learn to understand PDDL syntax, action semantics, state representations, and goal achievement from its general language understanding baseline. Phase 2 uses only 3 epochs because the model already possesses basic planning capabilities and needs only to refine its chain-of-thought reasoning processes. The reduced epoch count in Phase 2 also prevents overfitting to the specific feedback patterns generated by VAL, ensuring that the learned reasoning generalizes beyond the particular validation scenarios encountered during training.

\paragraph{Warmup Steps} Warmup steps implement a gradual increase in learning rate from zero to the target value at the beginning of training, preventing training instability that can arise from large initial weight updates on a partially trained model. Phase 1 uses 500 warmup steps to ensure stable convergence when adapting the pre-trained language model to the structured domain of planning, where the token distributions and semantic relationships differ significantly from general text. Phase 2 employs 200 warmup steps, fewer than Phase 1 because the model has already been adapted to the planning domain and requires less careful initialization. The warmup mechanism is particularly important in Phase 2 given the complex loss landscape created by the two-stage optimization process and the feedback-driven training dynamics.

\paragraph{Weight Decay} Weight decay implements L2 regularization by adding a penalty term proportional to the squared magnitude of model weights, preventing overfitting by discouraging the model from relying too heavily on specific parameter configurations. Phase 1 uses a weight decay of 0.01, relatively high to prevent the model from memorizing specific instruction-response patterns rather than learning generalizable planning principles. Phase 2 reduces weight decay to 0.001 to allow more fine-grained parameter adjustments necessary for learning subtle logical reasoning patterns while still providing some regularization against overfitting to the VAL feedback patterns. The lower weight decay in Phase 2 recognizes that the chain-of-thought reasoning requires precise parameter configurations that might be overly penalized by stronger regularization.

\paragraph{Gradient Clipping} Gradient clipping prevents exploding gradients by setting a maximum allowed norm for gradient vectors, ensuring training stability particularly in the complex optimization landscape of instruction tuning. Phase 1 employs gradient clipping at 1.0, providing stability during the initial adaptation from general language modeling to planning-specific tasks where gradient magnitudes can vary significantly across different types of planning problems. Phase 2 uses more conservative clipping at 0.5 because the model is more stable after Phase 1 training, and the chain-of-thought training process requires more careful weight updates to maintain the delicate balance between logical reasoning accuracy and plan generation quality. The tighter clipping in Phase 2 also helps manage gradient spikes that can occur when VAL feedback indicates dramatic plan validity changes.

\paragraph{Temperature (Generation)} The temperature parameter controls the randomness in text generation during training validation and inference, with lower values producing more deterministic outputs and higher values encouraging exploration of diverse response patterns. Phase 1 uses a temperature of 0.7, allowing moderate exploration of different planning approaches and explanation styles while maintaining coherent output quality. This higher temperature helps the model discover various ways to explain action validity and plan construction during the foundational learning phase. Phase 2 reduces temperature to 0.3 to focus generation on precise, logical reasoning steps where consistency and accuracy are more important than diversity. The lower temperature ensures that chain-of-thought reasoning follows logical patterns rather than exploring creative but potentially incorrect reasoning paths.

\paragraph{Maximum Generation Length} The maximum generation length sets the upper bound on tokens the model can produce in response to prompts, scaled to accommodate the verbosity requirements of each training phase. Phase 1 limits generation to 1024 tokens, sufficient for producing plans with basic explanations of action applicability and goal achievement without excessive computational cost. Phase 2 increases this limit to 2048 tokens to accommodate detailed step-by-step reasoning chains that include comprehensive state analysis, action justification, precondition verification, effect application reasoning, and goal progress tracking. This increased generation capacity is essential for the model to produce the verbose logical reasoning that characterizes effective planning verification and enables meaningful feedback from the VAL validator.

\paragraph{Optimizer (AdamW)} AdamW serves as the optimization algorithm for both training phases, chosen for its superior performance in transformer fine-tuning scenarios compared to standard optimizers. AdamW combines the adaptive learning rate benefits of Adam with improved weight decay handling, making it particularly effective for instruction tuning where the model must adapt pre-trained knowledge to new task-specific patterns. The optimizer handles sparse gradients well, which is crucial in planning scenarios where many potential actions are invalid in any given state, leading to sparse activation patterns. AdamW's momentum-based updates help navigate the complex loss landscape created by the combination of language modeling objectives and planning-specific constraints.

\paragraph{Beta Parameters ($\beta_1$, $\beta_2$)} The beta parameters control the exponential decay rates for AdamW's moment estimates, with $\beta_1 = 0.9$ governing the first moment (gradient moving average) and $\beta_2 = 0.999$ governing the second moment (squared gradient moving average). These standard values have proven effective across a wide range of transformer training scenarios and provide appropriate momentum characteristics for instruction tuning. The $\beta_1$ value of 0.9 provides sufficient momentum to smooth gradient noise while remaining responsive to genuine changes in gradient direction, particularly important when learning from VAL feedback in Phase 2. The $\beta_2$ value of 0.999 provides stable variance estimates essential for adaptive learning rate scaling across the diverse parameter space of large language models.

\paragraph{Epsilon ($\epsilon$)} The epsilon parameter adds a small constant of $1 \times 10^{-8}$ to the denominator in AdamW's update rule to prevent numerical instability from division by zero or near-zero values. This value represents a standard choice that provides numerical stability without meaningfully affecting the optimization dynamics. The parameter becomes particularly important during Phase 2 training where the complex loss landscape and feedback-driven updates can occasionally produce very small gradient variances that might otherwise cause numerical issues. The chosen value ensures robust training across the full range of planning problems and feedback scenarios encountered during instruction tuning.

\paragraph{Iteration Limit ($\eta$)} The iteration limit is unique to Phase 2 and controls how many feedback loops the model experiences with the VAL validator during chain-of-thought instruction tuning. Values of 10 and 15 represent the number of times the model can generate a plan with reasoning, receive detailed feedback about logical errors, learn from this feedback, and attempt improved solutions. This parameter directly controls the trade-off between training thoroughness and computational cost, as each iteration requires plan generation, validation, and model updating. Higher values of $\eta$ allow more refinement of reasoning capabilities but significantly increase training time and computational requirements. The specific values were chosen to provide sufficient learning opportunities while maintaining practical training times.

\subsection{Mathematical Formulation of Loss Functions}

We formally define the two specialized loss functions that drive our two-stage optimization process in Phase 2. These functions are carefully designed to target both the logical reasoning capabilities and final planning performance of the model.

\subsubsection{Reasoning Chain Loss Function}

The reasoning chain loss function $\mathcal{L}_{\text{reasoning}}$ measures the quality of the model's step-by-step logical reasoning over state-action-state transitions:

\begin{equation}
\mathcal{L}_{\text{reasoning}}(\theta_t, \mathbb{D}^t_{\text{reasoning}}) = \frac{1}{|\mathbb{D}^t_{\text{reasoning}}|} \sum_{(s_{i-1}, a_i, s_i, f_i) \in \mathbb{D}^t_{\text{reasoning}}} \mathcal{L}_{\text{step}}(s_{i-1}, a_i, s_i, f_i)
\end{equation}

where each training example consists of a state transition $(s_{i-1}, a_i, s_i)$ and VAL feedback $f_i$. The step-wise loss $\mathcal{L}_{\text{step}}$ is defined as:

\begin{equation}
\mathcal{L}_{\text{step}}(s_{i-1}, a_i, s_i, f_i) = d_{\text{state}}(s_i, s_i^{\text{expected}}) + \lambda_{\text{feedback}} \cdot \mathcal{L}_{\text{feedback}}(f_i)
\end{equation}

where $s_i^{\text{expected}}$ is the deterministically computed next state given action $a_i$ applied to $s_{i-1}$, and $d_{\text{state}}$ is the state distance function defined as:

\begin{equation}
d_{\text{state}}(s, s') = |s \triangle s'| = |s \setminus s'| + |s' \setminus s|
\end{equation}

This measures the symmetric difference between the two sets of predicates, counting predicates that are in one state but not the other. 

The feedback loss $\mathcal{L}_{\text{feedback}}$ incorporates VAL verification results to guide logical reasoning:

\begin{equation}
\mathcal{L}_{\text{feedback}}(f_i) = 
\begin{cases}
0 & \text{if action } a_i \text{ is valid} \\
\alpha_{\text{precond}} & \text{if precondition violation detected} \\
\alpha_{\text{effect}} & \text{if incorrect effect application} \\
\alpha_{\text{goal}} & \text{if goal achievement failure}
\end{cases}
\end{equation}

where $\alpha_{\text{precond}} = 1.0$, $\alpha_{\text{effect}} = 1.0$, $\alpha_{\text{goal}} = 1.5$ are penalty weights for different error types, and $\lambda_{\text{feedback}} = 0.1$ balances the feedback signal with the primary reasoning objective.

\subsubsection{Final Performance Loss Function}

The final performance loss function $\mathcal{L}_{\text{final}}$ measures how well the complete plans generated through chain-of-thought reasoning achieve the planning objectives:

\begin{equation}
\mathcal{L}_{\text{final}}(\theta_t^r, \mathbb{D}^t_{\text{final}}) = \frac{1}{|\mathbb{D}^t_{\text{final}}|} \sum_{(d, p, \pi, v) \in \mathbb{D}^t_{\text{final}}} \mathcal{L}_{\text{plan}}(d, p, \pi, v)
\end{equation}

where each training example consists of a domain $d$, problem $p$, generated plan $\pi$, and binary validity label $v$ from VAL. The plan-level loss is:

\begin{equation}
\mathcal{L}_{\text{plan}}(d, p, \pi, v) = \mathbb{I}[v = 0] \cdot \beta + \alpha \cdot \text{BCE}(v, \hat{v})
\end{equation}

where $\mathbb{I}[v = 0]$ is an indicator function that equals 1 when the plan is invalid (providing a fixed penalty $\beta = 2.0$ for invalid plans) and 0 when valid; and $\text{BCE}(v, \hat{v})$ is the binary cross-entropy loss between the VAL validity label $v$ and the model's predicted validity $\hat{v}$, with $\alpha = 0.5$ balancing plan generation accuracy with validity prediction.

\subsubsection{Dataset Construction for Loss Computation}

The reasoning dataset $\mathbb{D}^t_{\text{reasoning}}$ contains individual state-action-state triplets extracted from chain-of-thought sequences:
\begin{equation}
\mathbb{D}^t_{\text{reasoning}} = \{(s_{i-1}, a_i, s_i, f_i) : \forall \text{ steps in CoT plans generated at iteration } t\}
\end{equation}

The final dataset $\mathbb{D}^t_{\text{final}}$ contains complete planning instances with validity judgments:
\begin{equation}
\mathbb{D}^t_{\text{final}} = \{(d_j, p_j, \pi_j^t, v_j^t) : \forall \text{ problems } j \text{ at iteration } t\}
\end{equation}

where $\pi_j^t$ is the complete plan generated for problem $j$ at iteration $t$, and $v_j^t$ is the corresponding VAL validity assessment.

\newpage
\subsection{Algorithm}
\label{sec:preprocessing}

\begin{algorithm}
\caption{\name: Chain-of-Thought Instruction Tuning for Symbolic Planning}
\label{alg:pddl_instruct}
\begin{flushleft}
\textbf{Input}: Pre-trained LLM $M_{\theta_0}$, Phase 1 dataset $\mathbb{D}_1$, Phase 2 dataset $\mathbb{D}_2$, VAL validator,\\ \qquad \,\,\,\,\,\,iteration limit $\eta$, learning rates $\delta_1, \delta_2$\\
\textbf{Output}: Instruction-tuned model $M_{\theta^*}$
\end{flushleft}
\begin{algorithmic}[1]
\State \textbf{Phase 1: Initial Instruction Tuning}
\For{epoch $e = 1$ to $E_1$}
    \For{batch $(d_i, p_i, \pi_i, f_i) \in \mathbb{D}_1$}
        \State $y_i \leftarrow M_{\theta}(d_i, p_i)$ \Comment{Generate plan with explanation}
        \State $\mathcal{L}_1 \leftarrow -\log P(\pi_i, f_i | d_i, p_i, \theta)$
        \State $\theta \leftarrow \theta - \delta_1 \nabla_\theta \mathcal{L}_1$
    \EndFor
\EndFor
\State $\theta_1 \leftarrow \theta$ \Comment{Save Phase 1 model}

\State \textbf{Phase 2: CoT Instruction Tuning}
\For{iteration $t = 1$ to $\eta$}
    \State Initialize datasets $\mathbb{D}^t_{reasoning} \leftarrow \emptyset$, $\mathbb{D}^t_{final} \leftarrow \emptyset$
    
    \For{problem $(d_j, p_j) \in \mathbb{D}_2$}
        \State Generate CoT plan: $\pi_t^j = \{(s_0, a_1, s_1), (s_1, a_2, s_2), \ldots, (s_{n-1}, a_n, s_n)\}$
        \State using $M_{\theta_t}(d_j, p_j)$
        
        \State Validate plan with VAL: $f_j \leftarrow \text{VAL}(\pi_t^j, d_j, p_j)$
        
        \If{$f_j$ indicates valid plan}
            \State $\mathbb{D}^t_{final} \leftarrow \mathbb{D}^t_{final} \cup \{(d_j, p_j, \pi_t^j, 1)\}$
        \Else
            \State Extract detailed feedback for each invalid step
            \State $\mathbb{D}^t_{final} \leftarrow \mathbb{D}^t_{final} \cup \{(d_j, p_j, \pi_t^j, 0)\}$
        \EndIf
        
        \For{each step $(s_{i-1}, a_i, s_i) \in \pi_t^j$}
            \State Get step-level VAL feedback: $f_i \leftarrow \text{VAL-step}(s_{i-1}, a_i, s_i, d_j)$
            \State $\mathbb{D}^t_{reasoning} \leftarrow \mathbb{D}^t_{reasoning} \cup \{(s_{i-1}, a_i, s_i, f_i)\}$
        \EndFor
    \EndFor
    
    \State \textbf{Stage 1: Reasoning Chain Optimization}
    \For{epoch $e = 1$ to $E_{2a}$}
        \For{batch $B \in \mathbb{D}_t^{reasoning}$}
            \State $L_{reasoning} \leftarrow \frac{1}{|B|} \sum_{(s_{i-1}, a_i, s_i, f_i) \in B} L_{step}(s_{i-1}, a_i, s_i, f_i)$
            \State $\theta_t^r \leftarrow \theta_t - \delta_1 \nabla_{\theta_t} L_{reasoning}$
        \EndFor
    \EndFor
    
    \State \textbf{Stage 2: Final Performance Optimization}
    \For{epoch $e = 1$ to $E_{2b}$}
        \For{batch $B \in \mathbb{D}_t^{final}$}
            \State $\mathcal{L}_{final} \leftarrow \frac{1}{|B|} \sum_{(d, p, \pi, v) \in B} \mathcal{L}_{plan}(d, p, \pi, v)$
            \State $\theta_{t+1} \leftarrow \theta_t^r - \delta_2 \nabla_{\theta_t^r} \mathcal{L}_{final}$
        \EndFor
    \EndFor
\EndFor

\State \Return $M_{\theta^*}$ where $\theta^* = \theta_\eta$
\end{algorithmic}
\end{algorithm}

\newpage

\subsection{Hardware and Computational Resources}
\label{sec:compute}

\begin{table}[h!t]
\centering
\rowcolors{2}{gray!13}{}
\caption{Computational resource requirements}
\label{tab:compute_resources}
\begin{tabular}{lcc}
\toprule
\textbf{Resource} & \textbf{Phase 1} & \textbf{Phase 2} \\
\midrule
GPU Memory (per GPU) & 24 GB & 24 GB \\
Number of GPUs & 2 & 2 \\
Training Time & 12 hours & 18 hours \\
CPU Cores & 16 & 16 \\
RAM & 64 GB & 64 GB \\
\midrule
\textbf{Total Training Time} & \multicolumn{2}{c}{30 hours} \\
\textbf{Inference Time (per problem)} & \multicolumn{2}{c}{2.3 seconds} \\
\bottomrule
\end{tabular}
\end{table}

\section{Sample Prompts for Blocksworld Domain}
\label{sec:sample_prompts}

This section presents the specific prompt templates used in our \nameS framework for the Blocksworld domain. We provide examples for both Phase 1 (Initial Instruction Tuning) and Phase 2 (CoT Instruction Tuning) to demonstrate how our approach teaches models to reason about action applicability and state transitions.

\subsection{Phase 1: Initial Instruction Tuning Prompts}

\subsubsection{Correct Plan Example}

\mytcbinput{prompts/initial.tex}{Phase 1 Prompt - Correct Plan}{0}

\subsubsection{Incorrect Plan Example}

\mytcbinput{prompts/incorrect_initial.tex}{Phase 1 Prompt - Incorrect Plan}{0}

\subsection{Phase 2: Chain-of-Thought Instruction Tuning Prompts}

\subsubsection{CoT Generation Prompt}

\mytcbinput{prompts/cot_generation.tex}{Phase 2 CoT Generation Prompt}{0}

\subsubsection{CoT with Feedback Integration: Incorrect Plan}

\noindent\textbf{B.2.2.1 Binary Feedback} 

\vspace{0.05in}

\mytcbinput{prompts/cot_feedback_incorrect_binary.tex}{Phase 2 CoT Binary Feedback - Incorrect Plan}{0}

\noindent\textbf{B.2.2.2 Detailed Feedback} 

\vspace{0.05in}

\mytcbinput{prompts/cot_feedback_incorrect_detailed.tex}{Phase 2 CoT Detailed Feedback - Incorrect Plan}{0}

\subsubsection{CoT with Feedback Integration: Correct Plan}

\noindent\textbf{B.2.3.1 Binary Feedback} 

\vspace{0.05in}

\mytcbinput{prompts/cot_feedback_correct_binary.tex}{Phase 2 CoT Binary Feedback - Correct Plan}{0}

\noindent\textbf{B.2.3.2 Detailed Feedback} 

\vspace{0.05in}

\mytcbinput{prompts/cot_feedback_correct_detailed.tex}{Phase 2 CoT Detailed Feedback - Correct Plan}{0}

\section{Extended Experimental Results}

\subsection{Ablation Study Results}
\label{sec:ablation}

\begin{table}[h]
\centering
\rowcolors{2}{gray!13}{}
\caption{Ablation study showing contribution of each component for Llama-3}
\label{tab:ablation}
\begin{tabular}{lccc}
\toprule
\textbf{Configuration} & \textbf{Blocksworld} & \textbf{Mystery BW} & \textbf{Logistics} \\
\midrule
Baseline (No Training) & 28.0 $\pm$ 4.2 & 1.0 $\pm$ 1.0 & 11.0 $\pm$ 2.8 \\
Phase 1 Only & 78.0 $\pm$ 3.1 & 32.0 $\pm$ 4.6 & 23.0 $\pm$ 3.9 \\
Phase 2 Only (Detailed Feedback, $\eta=15$) & 72.0 $\pm$ 6.5 & 17.0 $\pm$ 3.2 & 45.0 $\pm$ 4.7 \\
Phase 1 + Binary Feedback ($\eta=15$) & 89.0 $\pm$ 2.7 & 49.0 $\pm$ 5.2 & 72.0 $\pm$ 4.1 \\
Phase 1 + Detailed Feedback ($\eta=15$) & \textbf{94.0 $\pm$ 1.5} & \textbf{64.0 $\pm$ 3.8} & \textbf{79.0 $\pm$ 3.2} \\
\bottomrule
\end{tabular}
\end{table}

\subsection{Error Analysis and Failure Modes}
\label{sec:error_analysis}

\begin{table}[h]
\centering
\rowcolors{2}{gray!13}{}
\caption{Breakdown of planning failures by error type (\%) for Llama-3 with Phase 1 and Phase 2 with Detailed Feedback and $\eta=15$}
\label{tab:error_breakdown}
\begin{tabular}{lccc}
\toprule
\textbf{Error Type} & \textbf{Blocksworld} & \textbf{Mystery BW} & \textbf{Logistics} \\
\midrule
Precondition Violation & 2.1 & 8.7 & 5.3 \\
Incorrect Effect Application & 1.4 & 12.4 & 6.8 \\
Goal Not Achieved & 1.8 & 9.2 & 6.1 \\
Invalid Action Sequence & 0.7 & 5.7 & 2.8 \\
\midrule
\textbf{Total Failure Rate} & \textbf{6.0} & \textbf{36.0} & \textbf{21.0} \\
\bottomrule
\end{tabular}
\end{table}

\end{document}